\documentclass[letterpaper]{article} % DO NOT CHANGE THIS
\usepackage{aaai2026}  % DO NOT CHANGE THIS
\usepackage{times}  % DO NOT CHANGE THIS
\usepackage{helvet}  % DO NOT CHANGE THIS
\usepackage{courier}  % DO NOT CHANGE THIS
\usepackage[hyphens]{url}  % DO NOT CHANGE THIS
\usepackage{graphicx} % DO NOT CHANGE THIS
\usepackage{times}
\usepackage{soul}
\usepackage{url}
\usepackage[utf8]{inputenc}
\usepackage{graphicx}
\usepackage{amsmath}
\usepackage{amsthm}
\usepackage{booktabs}
\usepackage{algorithm}
\usepackage{algorithmic}
\usepackage[switch]{lineno}
\usepackage{bm}
\usepackage{amsfonts,amssymb}
\usepackage{amsmath,amssymb,amsfonts}
\usepackage{tikz}  % 导入 tikz 包
\usepackage{algorithmic}
\usepackage{graphicx}
\usepackage{mathrsfs}
\usepackage{textcomp}
\usepackage{tikz}  
\usepackage{tikz} % 导入TikZ包
\usepackage{mathrsfs}
\usepackage{arydshln}
\usepackage{enumitem}
\usepackage{threeparttable}
\usepackage{array} 
\usepackage{booktabs,makecell, multirow, tabularx}
\usepackage[utf8]{inputenc}
\usepackage{xcolor}
\usepackage{tcolorbox}
\usepackage{booktabs}
\usepackage{array}
\usepackage{lipsum}
\usepackage{graphicx}  % 插入图片的宏包
\usepackage{subcaption}  % 用于子图的宏包
\usepackage{amsmath}
\usepackage{multicol} % 用于双栏排版
\usepackage{natbib}  % DO NOT CHANGE THIS AND DO NOT ADD ANY OPTIONS TO IT
\usepackage{caption} % DO NOT CHANGE THIS AND DO NOT ADD ANY OPTIONS TO IT
\usepackage{algorithm}
\usepackage{algorithmic}

\usepackage{newfloat}
\usepackage{listings}
\tcbset{
  promptstyle/.style={
    colback=blue!5!white,
    colframe=blue!75!black,
    fonttitle=\bfseries,
    coltitle=black,
    colbacktitle=blue!40!white,
    boxrule=0.5mm,
    sharp corners,
    boxsep=5pt,
    toptitle=2mm,
    bottomtitle=2mm
  },
  responsestyle/.style={
    colback=gray!10!white,      % 灰色背景色
    colframe=gray!50!black,     % 边框为深灰色
    fonttitle=\bfseries,
    coltitle=black,
    colbacktitle=gray!30!white, % 灰色标题背景
    boxrule=0.5mm,
    sharp corners,
    boxsep=5pt,
    toptitle=2mm,
    bottomtitle=2mm
  }
}
\newtcolorbox{promptbox}[1][]{title=Prompt,#1,promptstyle}
\newtcolorbox{responsebox}[1][]{title=Response,#1,responsestyle}

\newcommand{\sysname}{LECT}

\DeclareCaptionStyle{ruled}{labelfont=normalfont,labelsep=colon,strut=off} % DO NOT CHANGE THIS
\floatstyle{ruled}
\newfloat{listing}{tb}{lst}{}
\floatname{listing}{Listing}
\title{LLM-Enhanced Energy Contrastive Learning for Out-of-Distribution Detection in Text-Attributed Graphs}
\author{
    Xiaoxu Ma\textsuperscript{\rm 1}, 
    Dong Li\textsuperscript{\rm 2},
    Minglai Shao\textsuperscript{\rm 1,3}\thanks{Corresponding author.},
    Xintao Wu\textsuperscript{\rm 4},
    Chen Zhao\textsuperscript{\rm 2}
}
\affiliations{
    \textsuperscript{\rm 1}School of New Media and Communication, Tianjin University, China\\
    \textsuperscript{\rm 2}Department of Computer Science, Baylor University, USA\\
    \textsuperscript{\rm 3}Key Lab of Education Blockchain and Intelligent Technology, Ministry of Education, Guangxi Normal University,  China\\
    \textsuperscript{\rm 4}Electrical Engineering and Computer Science Department, University of Arkansas, USA\\
    \{maxiaoxu, shaoml\}@tju.edu.cn, \{dong\_li1, chen\_zhao\}@baylor.edu, xintaowu@uark.edu
}

\usepackage{bibentry}
\begin{document}    

\maketitle
\begin{abstract}
Text-attributed graphs, where nodes are enriched with textual attributes, have become a powerful tool for modeling real-world networks such as citation, social, and transaction networks. However, existing methods for learning from these graphs often assume that the distributions of training and testing data are consistent. This assumption leads to significant performance degradation when faced with out-of-distribution (OOD) data. In this paper, we address the challenge of node-level OOD detection in text-attributed graphs, with the goal of maintaining accurate node classification while simultaneously identifying OOD nodes. We propose a novel approach, \textbf{L}LM-Enhanced \textbf{E}nergy \textbf{C}ontrastive Learning for Out-of-Distribution Detection in \textbf{T}ext-Attributed Graphs (\textbf{LECT}), which integrates large language models (LLMs) and energy-based contrastive learning. The proposed method involves generating high-quality OOD samples by leveraging the semantic understanding and contextual knowledge of LLMs to create dependency-aware pseudo-OOD nodes, and applying contrastive learning based on energy functions to distinguish between in-distribution (IND) and OOD nodes. The effectiveness of our method is demonstrated through extensive experiments on six benchmark datasets, where our method consistently outperforms state-of-the-art baselines, achieving both high classification accuracy and robust OOD detection capabilities.
\end{abstract}

\section{Introduction}
    \label{sec:introduction}

Text-attributed graphs, which are a type of graph structure where nodes are associated with textual attributes, have gained significant attention due to their wide applicability in real-world scenarios, such as citation networks, social networks, e-commerce transaction networks, and hyperlink networks \cite{TAGbenchmark}. Current approaches to learning from text-attributed graphs \cite{TAGgraphformers,chen2021new,Tagexplanations,zhao2021fairness,wu2025explainable,lin2025face4fairshifts} typically follow a two-step process: first, extracting textual embeddings \cite{bert,roberta,minilm}, and then applying Graph Neural Networks (GNNs) \cite{GCN,GAT,Graphsage} for message passing. While these methods have achieved notable success, they rely on the assumption that the distribution of training and test data is consistent. In the presence of out-of-distribution (OOD) data, these methods often misclassify OOD nodes as part of existing categories, undermining the model’s robustness and reliability. Thus, detecting OOD anomalies while ensuring effective graph learning has become a critical challenge.

We address the problem of node-level OOD detection in text-attributed graphs, with a focus on achieving accurate node classification while effectively identifying OOD nodes. Unlike OOD detection in vision \cite{cvoodsurvey,shao2024supervised} and language \cite{nlpoodsurvey}, where OOD samples are assumed to be independent and identically distributed (i.i.d.), OOD detection in graph data is significantly more complex due to the non-Euclidean nature of graphs and the interdependence of their data. Existing methods for graph OOD detection \cite{graphoodsurvey} can be categorized into propagation-based and classification-based approaches. Propagation-based methods \cite{gpn,oodgat,ossnc} leverage the message-passing paradigm of GNNs to propagate uncertainty estimates, while classification-based methods \cite{msp,odin,graphsafe} design OOD score metrics for detection. However, most current approaches overlook textual attributes and rely on shallow models, such as bag-of-words, which may fail to capture the rich semantic information embedded in the text. Furthermore, these models often focus on complex propagation mechanisms and uncertainty quantification, while neglecting the critical interplay between textual features and node connectivity. With the rapid advancement of large language models (LLMs) \cite{li2025solverllm,li2025multi,zhao2025uncertainty}, which provide extensive contextual knowledge and semantic understanding, textual attribute information can now be more effectively captured and leveraged for OOD detection \cite{cv_envisioning,li2024learning,llmoodsurvey}.

To address OOD detection in text-attributed graphs, we propose a novel LLM-enhanced contrastive learning method, named \textbf{LECT}. Our method consists of two key stages: (1) LLM-based sample generation and (2) energy-based contrastive learning. In the first stage, we generate pseudo-OOD nodes by randomly initializing edges. Using the textual understanding and generation capabilities of LLMs, we create OOD samples related to connected IND nodes through chain-of-thought (COT) prompts, producing high-quality pseudo-data with meaningful dependency relationships. In the second stage, we train the model using Linked IND-OOD Pairs and Triplet Contrastive Pairs, applying an energy-based objective function to effectively distinguish between IND and OOD samples. Our contributions are as follows:
\begin{itemize}[leftmargin=*]
    \item We propose a novel method for generating high-quality OOD node textual attributes using LLMs, ensuring dependency relationships with connected IND nodes.
    \item We design an energy-based contrastive learning algorithm for OOD detection, which enables multi-dimensional sample pairs to effectively capture distinctions between IND and OOD samples.
    \item We demonstrate that our model outperforms state-of-the-art baselines on six benchmark datasets while maintaining strong node classification performance.
\end{itemize}

\section{Related Work}
    \label{sec:related_work}

\textbf{LLM-based Graph Mining}. Large language models (LLMs) have shown strong capabilities in text representation and generation, leading to their broad adoption in graph learning tasks \cite{llmgraphsurvey,llmsurvey2}. Their roles can be categorized as enhancers \cite{tape,Giant,OFAenhancer}, predictors \cite{graphgpt,llaga,graphtext}, and aligners \cite{graphtext,GLEM}. As enhancers, LLMs enrich node features (e.g., TAPE \cite{tape}); as predictors, they perform reasoning-based tasks (e.g., GraphGPT \cite{graphgpt}); and as aligners, they ensure consistency between language and graph modalities (e.g., GLEM \cite{GLEM}). Among these, using LLMs to enhance node attributes is notably efficient and stable \cite{glbench}.

While prior work has focused on node classification and link prediction, little attention has been given to OOD detection in text-attributed graphs. Motivated by the enhancer role, we propose leveraging LLMs to improve node-level OOD detection by generating diverse textual attributes and extracting high-quality representations from pretrained models.

\textbf{OOD Detection on Graphs.} Despite the strong representational capabilities of GNN, their performance often degrades or becomes overconfident when test samples differ significantly from the training distribution, leading to misclassification. Consequently, OOD detection on graphs has attracted considerable attention and can be categorized into graph-level and node-level tasks \cite{graphoodsurvey,li2023contrastive}. Node-level OOD detection focuses on OOD nodes within a single graph. Due to the interdependence among nodes, it presents greater challenges compared to graph-level detection \cite{graphood,graphood2}, which involves separate graphs.

Currently, node-level OOD detection methods can be broadly divided into classification-based and propagation-based approaches. Classification-based methods, such as MSP \cite{msp} and ODIN \cite{odin}, utilize maximum softmax probability or temperature scaling to estimate OOD probabilities. However, these methods struggle to handle complex distributions and are prone to overconfidence issues. Propagation-based methods, such as GPN \cite{gpn} and OODGAT \cite{oodgat}, enhance model reliability by propagating uncertainty within the graph. GraphSAFE \cite{graphsafe} achieves promising results using energy-based techniques. However, it overlooks node textual attributes, and the scarcity of OOD samples during training, along with the complexity of setting upper and lower thresholds in real-world scenarios, limits its ability to effectively address complex node dependencies and distribution shifts.

To tackle these challenges, we propose a novel approach that leverages the generative and reasoning capabilities of LLMs to capture textual attribute dependencies among nodes, generating pseudo-OOD samples. Subsequently, energy-based contrastive learning is employed to capture the relationship between IND and OOD samples.

\section{Preliminaries}
    \label{sec:preliminaries}
    \begin{figure*}[h!] % 使用 figure* 来跨越两栏
    \centering
    \includegraphics[width=0.94\textwidth]{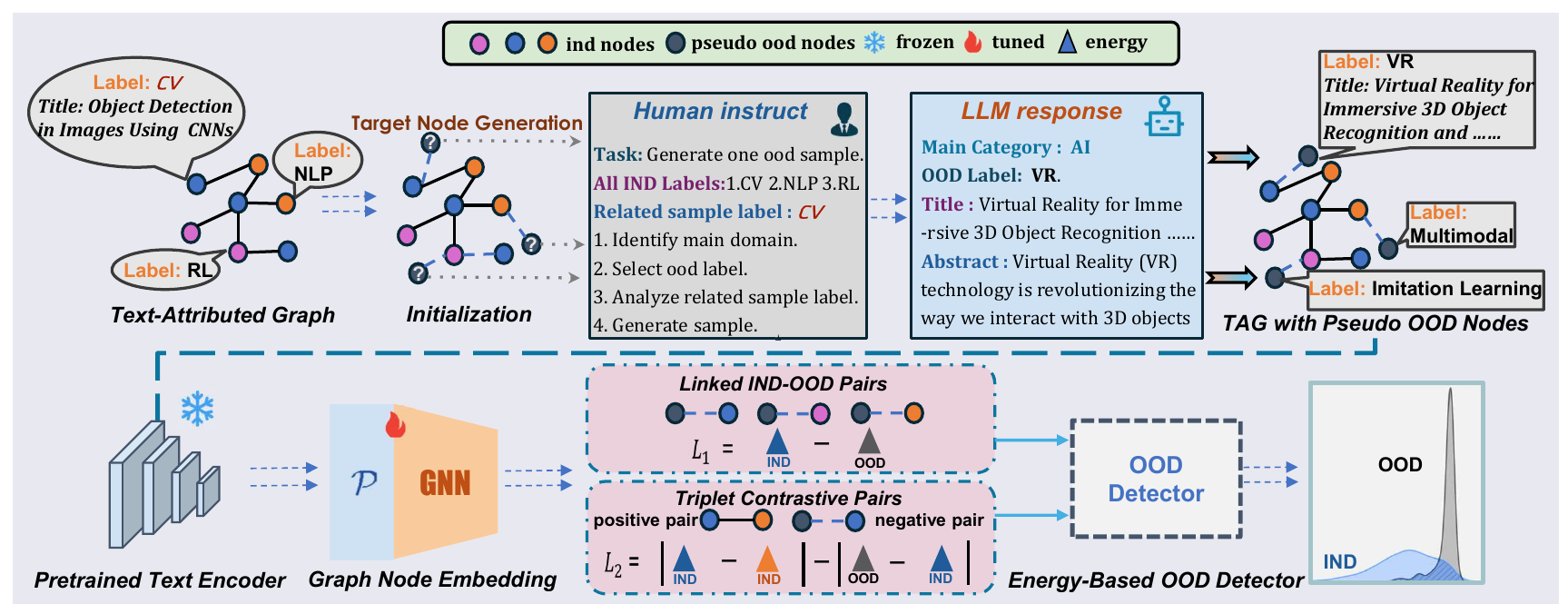} % 使用页面宽度填充
    \caption{ 
The overall pipeline of \sysname{}. Given a text-attributed graph, we first construct pseudo-OOD nodes by injecting random edges and generating their textual content with an LLM. We then derive node representations using a text encoder and a GNN with a projector. Energy scores are subsequently computed to form Linked IND–OOD Pairs and Triplet Contrastive Pairs for training. Finally, the model identifies OOD samples based on the resulting energy levels.}
    %\vspace{-5pt} % 控制图像和下方文本的间距
    \label{fig:model}
\end{figure*}

The node-level OOD detection task aims to determine whether each node in a graph belongs to the in-distribution or out-of-distribution. Specifically, given a graph with textual features, the goal is to classify each node based on its textual features and the graph's structural information, thereby identifying whether the node belongs to the training dataset's distribution or to an OOD distribution.

\textbf{Predictive Tasks on Text-Attributed Graphs.} Consider a text-attributed graph \( \mathbf{G} = (\mathbf{V}, \mathbf{E}, \mathbf{Q}) \), where \( \mathbf{V} \) is the set of nodes, \( \mathbf{E} = \{e_{ij}\} \) represents the set of edges, and the adjacency matrix \( \mathbf{A} \in \mathbb{R}^{|\mathbf{V}| \times |\mathbf{V}|} \) can be derived from the edge set. The adjacency matrix \( a_{ij} = 1 \) if there is an edge connecting nodes \( i \) and \( j \), and \( a_{ij} = 0 \) otherwise. Each node \( n \in \mathbf{V} \) is associated with a textual feature \( q_n \), typically a sentence or description. The set \( \mathbf{Q} = \{q_n\}_{n \in \mathbf{V}} \) denotes the collection of these textual features for all nodes. The node classification problem is formulated as follows: given labeled nodes \( \mathbf{Y}_{train} \) (where \( \mathbf{V}_{train} \subset \mathbf{V} \)), the goal is to predict the labels of the unlabeled nodes \( \mathbf{Y}_{test} \) (where \( \mathbf{V}_{test} = \mathbf{V} \setminus \mathbf{V}_{train} \)).

\textbf{OOD Detection Task on Text-Attributed Graphs.} Besides achieving strong predictive performance on IND nodes, the learned classifier is also expected to detect OOD instances that originate from a distinct data-generating distribution. For the node-level OOD detection task, we assume a training dataset \( \mathbf{V}_{train} = \{ V_{train}^1, V_{train}^2, \dots, V_{train}^{N_1} \} \) consisting solely of IND nodes, and a testing dataset \( \mathbf{V}_{test} = \{ V_{test_{in}}^1, V_{test_{in}}^2, V_{test_{out}}^1, \dots, V_{test_{out}}^{N_2} \} \), where \( V_{test_{in}} \) and \( V_{test_{out}} \) are drawn from distinct distributions \( P_{in} \) and \( P_{out} \), respectively. For each node \( V_{test}^n \), the goal of OOD detection is to determine, based on its textual feature \( Q_n \) and the graph's structural information \( \mathbf{A} \), whether the node originates from the training data distribution \( P_{in} \) or from an OOD distribution \( P_{out} \), while ensuring accurate node classification. 

We define a detector \( d \) and a scoring function \( s(V_{test}^n, f_\theta) \), where \( f_\theta \) represents the model trained on \( \mathbf{V}_{train} \) with parameters \( \theta \). The OOD detection task can be formalized as:
\begin{equation}
\small
d(V_{test}^n; \tau, s, f_\theta) =
\begin{cases}
V_{test}^n \in P_{in}, & \text{if } s(V_{test}^n, f_\theta) \leq \tau, \\
V_{test}^n \in P_{out}, & \text{if } s(V_{test}^n, f_\theta) > \tau,
\end{cases}    
\end{equation}
where \( s(V_{test}^n, f_\theta) \) is the scoring function that computes the score of node \( V_{test}^n \) based on its textual features and graph structure. The threshold \( \tau \) determines whether the node belongs to the in-distribution or out-of-distribution. \( P_{in} \) denotes the distribution of the training data, while \( P_{out} \) represents the OOD data distribution. The optimization goal of OOD detection is to learn an appropriate scoring function \( s(V_{test}^n, f_\theta) \) and well-trained model parameters \( f_\theta \).

\section{Methodology}
    \label{sec:methodology}
    In this section, we introduce the proposed \sysname{} algorithm for text-attributed graph OOD detection. The architecture of our proposed model is shown in Figure \ref{fig:model}.
%In Section \ref{sec:oodSampleGeneration}, we discuss how to generate OOD samples using large models, while in Section \ref{sec:energyContrastive}, we present the energy-based contrastive learning algorithm. 

\subsection{Pseudo-OOD Sample Generation}
    \label{sec:oodSampleGeneration}

\textbf{Initialization of Pseudo-OOD Nodes.}
Assume we have the original training graph \( \mathbf{G}_s = (\mathbf{V}_s, \mathbf{E}_s, \mathbf{Q_s}) \). First, we initialize the pseudo-OOD node set, resulting in \( \mathbf{V}_o = \{ v_o^1, v_o^2, \dots, v_o^{N_o} \} \), where \( N_o \) is the number of generated pseudo-OOD nodes.

To simulate the generation of OOD samples while modeling the dependencies between IND and OOD samples, we initialize the structural relationships between nodes using a random edge connection method. Specifically, each pseudo-OOD node \( v_o \) is randomly connected to one or more IND nodes in the graph. The number of IND nodes connected to each OOD node is controlled by a predefined upper limit \( C_{\text{max}} \), which means that each pseudo-OOD node can connect to a maximum of \( C_{\text{max}} \) IND nodes. This step aims to control the complexity of the graph, preventing excessive connections from introducing noise and increasing the inference difficulty for the subsequent large model. To control complexity, we initialize \( C_{\text{max}} \) as the number of classes in the training samples, that is, \( |Y_{\text{train}}| \).

Thus, we define the edge set \( \mathbf{E}_o \) as:
\begin{equation}
 \mathbf{E}_o = \{ e_{ij} = (v_i, v_j) \mid v_i \in \mathbf{V}_s, v_j \in\mathbf{V}_o \}.   
\end{equation}

\textbf{Chain-of-Thought (COT) Construction.}  
Given the initialized pseudo-node structure \( \mathbf{V}_{\text{o}} \) and edge set \( \mathbf{E}_{\text{o}} \), we employ LLM to generate OOD text attributes \(\mathbf{Q}_{\text{o}} \) based on the dependency relationships of the nodes. To address potential challenges in generating accurate and coherent responses for unfamiliar patterns, we incorporate the COT technique to guide step-by-step reasoning. This enables the model to produce high-quality OOD samples through logical and structured reasoning. The COT-guided generation process is as follows:

\begin{enumerate}[leftmargin=*]
    \item \textit{\textbf{Main Domain Classification:}} Using the labels of all IND samples, the LLM classifies the samples into known domains, aiding in categorizing OOD samples based on existing IND labels.
    
    \item \textit{\textbf{Selection of OOD Categories:}} The LLM selects an OOD category significantly different from the IND distribution to ensure the generated samples effectively represent out-of-distribution data.
    
    \item \textit{\textbf{Analysis of Connected Sample Labels:}} To simplify reasoning, only the labels of IND samples connected to the OOD nodes are provided. This allows the LLM to select OOD labels correlated with connected IND labels, simulating real-world graph dependencies.
    
    \item \textit{\textbf{Sample Generation:}} Based on the selected OOD label, the LLM generates textual features for the pseudo-OOD nodes.
\end{enumerate}

To simulate diverse OOD scenarios, the COT approach generates both near-OOD and far-OOD samples:  
\begin{itemize}[leftmargin=*]
    \item \textit{Near-OOD samples}: OOD categories are selected close to the main IND domain to capture subtle distributional shifts.  
    \item \textit{Far-OOD samples}: Categories are chosen farther from the IND domain to introduce more diverse and loosely related samples.  
\end{itemize}

For each pseudo-OOD node, we repeat the above LLM generation step, resulting in the pseudo-OOD node text features \( \mathbf{Q}_{\text{o}} \). Thus, the LLM-enhanced graph can be represented as:
\begin{equation}
 \mathbf{G}_{\text{en}} = \left( \mathbf{V}_s \cup \mathbf{V}_o, \mathbf{E}_s \cup \mathbf{E}_o, \mathbf{Q}_s \cup \mathbf{Q}_o \right). 
 \label{eq:Gen}
\end{equation}

\subsection{Energy-based Contrastive Learning}
    \label{sec:energyContrastive}   
\textbf{Text-Attributed Graphs Feature Extraction.} After generating the pseudo-OOD samples and obtaining the enhanced graph \( \mathbf{G}_{\text{en}} \), we use a pre-trained language model (PLM) to extract the textual features for each node. Specifically, for each node \( v_i \in \mathbf{G}_{\text{en}} \), the textual feature \( q^i \) is mapped to a textual embedding \( h^i \) via the PLM:
\begin{equation}
 h^i = \text{PLM}(q^i), \quad \forall i \in V_{\text{en}}.
\end{equation}

To align the textual features with the graph structure and capture richer representations, we input all node embeddings \( \mathbf{H} = [h^1, h^2, \dots, h^{|V_{\text{en}}|}] \) into a projector \( \mathcal{P} \), obtaining the projected features \( \mathbf{H'} \):
\begin{equation}
\mathbf{H'} = \mathcal{P}(\mathbf{H}).
\end{equation}

We then pass the projected textual features \( \mathbf{H'} \) along with the graph structure information \( \mathbf{A} \) into a GNN for message passing and node updating, resulting in the final node representations \( z^i \):
\begin{equation}
z^i = \text{GNN}(\mathbf{H'},\mathbf{A}), \quad \forall i \in \mathbf{V}_{\text{en}}.
\label{eq:GNN}
\end{equation}

Thus, the node representations \(\mathbf{Z} = [z^1, z^2, \dots, z^{|\mathbf{V}_{\text{en}}|}] \) are obtained for each node in the graph \( \mathbf{G}_{\text{en}} \).

\textbf{Energy-Based OOD Detection.} We use the energy function as the OOD detection score, which is grounded in the principles of Energy-Based Models (EBM) \cite{energycvood}. Specifically, the energy function quantifies the degree of alignment between a sample and the training data. The relationship between the energy function \( \mathcal{E}(x, y) \) and probability is described by the Boltzmann distribution, which can be written as:
\begin{equation}
 p(y | x) = \frac{e^{-\mathcal{E}(x, y)}}{\mathcal{Z}(x)} = \frac{e^{-\mathcal{E}(x, y)}}{\sum_{y'} e^{-\mathcal{E}(x, y')}},
 \label{eq:energy_logits}
\end{equation}
where \( \mathcal{E}(x, y) \) represents the energy for input \( x \) and class label \( y \), and \( \mathcal{Z}(x) \) is the partition function, which sums over all possible class labels. A lower energy \( \mathcal{E}(x, y) \) indicates a higher degree of alignment between sample \( x \) and class label \( y \), suggesting that the sample is more likely to belong to the training data distribution (i.e., IND samples). Hence, IND samples have lower energy values, while OOD samples exhibit higher energy values.

For the embedding representations \( z_i \) obtained from the PLM and graph neural network, we calculate the energy for each node using the following formula:

\begin{equation}
\mathcal{E}_i = - \log \left( \sum_{c=1}^{C} \exp(z^i_c) \right),
\label{eq:energy}
\end{equation}
where \( C \) is the number of classes, and \( z^i_c \) denotes the score for node \( v_i \) belonging to class \( c \).

\textbf{Linked IND-OOD Pairs.} We first construct pairs of linked IND and OOD samples. Using the energy values \( \mathbf{G}_{\text{en}} \) from Equation \ref{eq:Gen}, we randomly select a pair \( e_{ij} \) from the set \( E_o \), where \( v_i \) is an IND sample and \( v_j \) is an OOD sample, forming a linked IND-OOD sample pair \( \mathbb{P}_{\text{ind-ood}} \). Then, applying the vector \( z^i \) from Equation \ref{eq:GNN} and using Equation \ref{eq:energy}, we compute the corresponding energy values \( \mathcal{E}_i \) and \( \mathcal{E}_j \) for the IND and OOD samples, respectively.

For all IND-OOD pairs \( \mathbb{P}_{\text{ind-ood}} \), the loss is computed as:
\begin{equation}
\mathcal{L}_{\text{ind-ood}} = \mathbb{E}_{(v_i,v_j) \in \mathbb{P}_{\text{ind-ood}}} \left[ \max \left( 0, \gamma - (\mathcal{E}_j - \mathcal{E}_i) \right) \right],
\label{eq:ind-ood}
\end{equation}
where \( \mathcal{E}_i \) and \( \mathcal{E}_j \) are the energy values for the IND and OOD samples, respectively. \( \mathbb{P}_{\text{ind-ood}} \) denotes the set of all IND and OOD sample pairs, and \( (v_i,v_j) \) represents a specific pair. The energy difference \( \mathcal{E}_j - \mathcal{E}_i \) captures the dependency between the IND and OOD samples. The margin \( \gamma \) is a hyperparameter that enforces a lower bound on the energy difference, preventing misclassification of challenging pairs as OOD. According to Equation \ref{eq:energy_logits}, larger energy values indicate lower confidence, increasing the likelihood that a sample belongs to the OOD class. This loss function aims to minimize the energy difference to correctly distinguish between IND and OOD samples. Additionally, a constraint term is introduced to regulate the energy difference between the mean energies of all IND and OOD samples.

\textbf{Triplet Contrastive Pairs.} In this step, we construct triplet contrastive node pairs. Specifically, for a given IND node \( v_{\text{c}} \) in \( \textbf{G}_{\text{en}} \), we first identify the edge \( e_{ci} = (v_c, v_i) \) in \( \textbf{E}_s \), where \( v_i \) is an IND node, and the edge \( e_{cj} = (v_c, v_j) \) in \( \textbf{E}_o \), where \( v_j \) is an OOD node. The pair \( e_{ci} \) is treated as a positive sample, and \( e_{cj} \) as a negative sample, thus forming the triplet contrastive pair \( \mathcal{T} = \{ v_i, v_c, v_j \} \), where \( v_i \) and \( v_c \) are IND nodes, and \( v_j \) is an OOD node. The energy values for each node are then computed using Equation \ref{eq:GNN} and Equation \ref{eq:energy}.

The contrastive learning loss is defined as follows:
\begin{equation}
 \mathcal{L}_{\text{triplet}} = \mathbb{E}_{v_i, v_j, v_c \in \mathcal{T}} \left[ \max \left( 0, \left| \mathcal{E}_i - \mathcal{E}_{\text{c}} \right| - \left( \mathcal{E}_j - \mathcal{E}_{\text{c}} \right) \right) \right],
 \label{eq:triplet}
\end{equation}
where \( \left| \mathcal{E}_i - \mathcal{E}_{\text{c}} \right| \) represents the absolute energy difference between the positive sample nodes, and \( \mathcal{E}_j - \mathcal{E}_{\text{c}} \) represents the energy difference between the negative sample nodes.

The design of this loss function aims to capture the relationships and dependencies between IND and OOD nodes in the graph effectively. For the positive sample pair (the relationship between the IND node and the center node), we expect the energy difference to be as small as possible, since connected IND nodes should exhibit high similarity, implying a small energy difference. For the negative sample pair (the relationship between the OOD node and the center node), we expect a larger energy difference. OOD nodes typically exhibit lower confidence than IND nodes, so their energy values should be significantly higher, leading to a more pronounced energy difference.

% This design ensures that the model learns to distinguish between normal dependencies (i.e., the relationship between IND nodes) and abnormal dependencies (i.e., the relationship between OOD nodes and IND nodes).

% \textbf{Model Learning.} The main task of the model is node classification, optimized using a supervised loss function \( \mathcal{L}_{\text{sup}} \). This is formulated as a cross-entropy loss over the labeled nodes \( \textbf{V}_s \) in the training set. Let \( z_i \) denote the embedding vector of each node \( v_i \), the cross-entropy loss is defined as:
% \begin{equation}
%  \small
%  \mathcal{L}_{\text{sup}} = - \sum_{v \in \textbf{V}_s} \left[ y_v \log(\hat{y}_v) + (1 - y_v) \log(1 - \hat{y}_v) \right],   
%  \label{eq:sup}   
% \end{equation}

% where \( y_v \) is the true label of node \( v \), and \( \hat{y}_v \) represents the predicted probability of the corresponding label.
\textbf{Model Learning.} The main task of the model is node classification, optimized using a supervised loss function \( \mathcal{L}_{\text{sup}} \). This is formulated as a cross-entropy loss over the labeled nodes \( \mathbf{V}_s \) in the training set. Let \( \hat{\mathbf{y}}_v = (\hat{y}_{v,1}, \hat{y}_{v,2}, \ldots, \hat{y}_{v,C}) \) denote the predicted class probabilities for node \( v \), and \( y_v \in \{1, \ldots, C\} \) be the true class label. The cross-entropy loss is defined as:

\begin{equation}
\mathcal{L}_{\text{sup}} = - \sum_{v \in \mathbf{V}_s} \log \hat{y}_{v,y_v},
\label{eq:sup}
\end{equation}where \( \hat{y}_{v,y_v} \) is the predicted probability of the true class \( y_v \) for node \( v \).

Combining the supervised loss, IND-OOD loss, and triplet contrastive loss, the overall loss function \( \mathcal{L}_{\text{all}} \) is formulated as:
\begin{equation}
\mathcal{L}_{\text{all}} = \mathcal{L}_{\text{sup}} + \lambda_1 \mathcal{L}_{\text{ind-ood}} + \lambda_2 \mathcal{L}_{\text{triplet}},   
\end{equation}
where \( \mathcal{L}_{\text{ind-ood}} \) accounts for the pairwise loss between IND and OOD samples, and \( \mathcal{L}_{\text{triplet}} \) is the triplet contrastive loss. The hyperparameters \( \lambda_1 \) and \( \lambda_2 \) control the contributions of these auxiliary losses.

This combined loss enables the model to optimize for both node classification and OOD detection. %Theoretical justification for this design is provided in Appendix~D.
During testing, node embeddings are extracted and their energy values are computed to distinguish IND and OOD samples via a threshold. To avoid overfitting to pseudo-OOD samples, we freeze the pretrained language model (PLM) and update only the projection and GNN parameters. This reduces training overhead while preserving the PLM's generalization ability.

% During testing, we use the trained model to obtain node embeddings and compute energy values, applying a threshold to distinguish between IND and OOD samples.Notably, PLMs (pretrained language models) inherently generate high-quality text representations, reducing the need for extensive fine-tuning. However, excessive adaptation to pseudo-OOD samples may hurt generalization. To mitigate this, we freeze the PLM parameters and optimize only the projection and GNN parameters during training. This approach not only enhances training efficiency but also ensures strong generalization by minimizing the number of parameters that need to be updated.

\section{Experiments}
    \label{sec:experiments}

\begin{table*}[htbp]
\centering

% --- 把表格全部缩放到版心宽度 ---
\resizebox{\textwidth}{!}{%
\begin{threeparttable}

\scriptsize
\setlength{\abovecaptionskip}{0cm}
\renewcommand\arraystretch{0.9}
\setlength{\tabcolsep}{3pt}

\begin{tabular}{lcccccccccccc}
\toprule
 & \multicolumn{4}{c}{\texttt{Cora}} & \multicolumn{4}{c}{\texttt{Citeseer}} & \multicolumn{4}{c}{\texttt{Pubmed}} \\
\cmidrule(lr){2-5} \cmidrule(lr){6-9} \cmidrule(lr){10-13}
 & IND-Acc$\uparrow$ & AUROC$\uparrow$ & AUPR$\uparrow$ & FPR95$\downarrow$
 & IND-Acc$\uparrow$ & AUROC$\uparrow$ & AUPR$\uparrow$ & FPR95$\downarrow$
 & IND-Acc$\uparrow$ & AUROC$\uparrow$ & AUPR$\uparrow$ & FPR95$\downarrow$ \\
\midrule
      MSP & 86.72\tiny$\pm$0.47 & 81.96\tiny$\pm$1.38 & 43.37\tiny$\pm$1.23 & 99.21\tiny$\pm$0.43 
      & 75.36\tiny$\pm$0.76 & 71.47\tiny$\pm$1.29 & 30.81\tiny$\pm$0.57 & 97.64\tiny$\pm$0.89 
      &\textbf{92.68}\tiny$\pm$\textbf{0.18} & 60.53\tiny$\pm$3.38 & 48.67\tiny$\pm$3.71 & 96.65\tiny$\pm$0.94 \\
  
      ODIN & 86.78\tiny$\pm$0.21 & 80.76\tiny$\pm$1.96 & 43.92\tiny$\pm$4.48 & 62.74\tiny$\pm$11.61 
      & 76.40\tiny$\pm$0.50 & 70.60\tiny$\pm$2.27 & 32.97\tiny$\pm$3.78 & 79.77\tiny$\pm$3.09 
      & 92.26\tiny$\pm$0.29 & 59.86\tiny$\pm$3.00 & 49.09\tiny$\pm$4.07 & 92.39\tiny$\pm$1.32 \\

      OSSNC & 86.22\tiny$\pm$0.41 & 82.32\tiny$\pm$1.76 & 47.01\tiny$\pm$4.34 & 97.51\tiny$\pm$0.63 
      & 75.89\tiny$\pm$0.40 & 72.14\tiny$\pm$1.57 & 33.27\tiny$\pm$1.67 & 96.23\tiny$\pm$2.44 
      & \underline{92.50\tiny$\pm$0.28} & 61.14\tiny$\pm$2.32 & 48.84\tiny$\pm$2.38 & 97.22\tiny$\pm$0.36 \\
      
      EMP & 86.14\tiny$\pm$0.58 & 82.12\tiny$\pm$2.49 & 47.00\tiny$\pm$5.30 & 60.91\tiny$\pm$4.58 
      & 75.89\tiny$\pm$0.51 & 70.16\tiny$\pm$2.06 & 30.87\tiny$\pm$2.27 & 79.26\tiny$\pm$2.46 
      & 92.59\tiny$\pm$0.31 & 61.44\tiny$\pm$4.26 & 51.24\tiny$\pm$4.63 & 92.90\tiny$\pm$1.86 \\

      OODGAT & 88.08\tiny$\pm$0.65 & 80.31\tiny$\pm$1.47 & 37.51\tiny$\pm$4.25 & 49.09\tiny$\pm$5.82 
      &\underline{76.90\tiny$\pm$0.65} & 68.43\tiny$\pm$1.26 & 27.33\tiny$\pm$2.49 & 97.72\tiny$\pm$2.04
      & 91.52\tiny$\pm$0.26 & 70.08\tiny$\pm$1.39 & 32.67\tiny$\pm$3.99 & 96.55\tiny$\pm$1.78 \\

      GPN & 86.84\tiny$\pm$1.16 & 79.71\tiny$\pm$4.50 & 45.33\tiny$\pm$5.00 & 76.66\tiny$\pm$4.46 
      &76.76\tiny$\pm$0.38 & 55.35\tiny$\pm$1.05 & 46.38\tiny$\pm$0.61 & 92.51\tiny$\pm$1.17
      & 92.03\tiny$\pm$0.23 & 60.72\tiny$\pm$1.94 & 49.04\tiny$\pm$2.79 & 95.80\tiny$\pm$1.97 \\

        NODESAFE & 88.05\tiny$\pm$0.41 & 92.54\tiny$\pm$0.77 & 71.74\tiny$\pm$2.37 & 37.59\tiny$\pm$3.21 
         & 75.89\tiny$\pm$0.53 & 84.29\tiny$\pm$0.93 & 48.41\tiny$\pm$0.73 & \underline{52.92\tiny$\pm$2.37}
         & 91.42\tiny$\pm$0.33 & 83.60\tiny$\pm$1.38 & 79.34\tiny$\pm$3.01 & 63.92\tiny$\pm$4.77 \\

GRASP & 87.81\tiny$\pm$0.43 & 92.71\tiny$\pm$0.82 & 72.04\tiny$\pm$3.18 & 31.40\tiny$\pm$3.31 
      & 76.52\tiny$\pm$0.39 & 82.01\tiny$\pm$0.87 & 54.79\tiny$\pm$0.66 & 71.45\tiny$\pm$5.37
      & 90.39\tiny$\pm$0.77 & 83.79\tiny$\pm$0.93 & 79.77\tiny$\pm$1.88 & 72.17\tiny$\pm$5.21 \\

      GNNSAFE & 88.13\tiny$\pm$0.37 & 92.23\tiny$\pm$0.90 & 70.09\tiny$\pm$3.50 & 34.49\tiny$\pm$4.35 
      &76.70\tiny$\pm$0.62 & 81.20\tiny$\pm$1.31 & 51.17\tiny$\pm$3.65 & 69.17\tiny$\pm$2.47
      &90.72\tiny$\pm$0.46 & 82.55\tiny$\pm$1.22 & 75.67\tiny$\pm$2.80 & 71.26\tiny$\pm$6.03 \\ 

      GNNSAFE++ & 87.26\tiny$\pm$0.63 & 93.03\tiny$\pm$0.50 & 
      71.37\tiny$\pm$2.64 & 30.77\tiny$\pm$5.52 
      &73.76\tiny$\pm$2.30 & 84.51\tiny$\pm$1.98 & 49.23\tiny$\pm$10.93 & 54.77\tiny$\pm$6.29
      &81.89\tiny$\pm$2.59 & 84.62\tiny$\pm$2.25 & 70.76\tiny$\pm$5.77 & 58.98\tiny$\pm$2.33 \\ 
      
    %   \midrule
    %     LLaMA3 & 74.30\tiny$\pm$0.63 & 62.38\tiny$\pm$0.50 & 
    %   21.03\tiny$\pm$2.64 & 81.94\tiny$\pm$5.52 
    %   &73.76\tiny$\pm$2.30 & 84.51\tiny$\pm$1.98 & 49.23\tiny$\pm$10.93 & \underline{54.77\tiny$\pm$6.29}
    %   &81.89\tiny$\pm$2.59 & 84.62\tiny$\pm$2.25 & 70.76\tiny$\pm$5.77 & 58.98\tiny$\pm$2.33 \\

    %     Qwen2.5 & 66.96\tiny$\pm$0.63 & 51.90\tiny$\pm$0.50 & 
    %   59.58\tiny$\pm$2.64 & 19.49\tiny$\pm$5.52 
    %   &73.76\tiny$\pm$2.30 & 84.51\tiny$\pm$1.98 & 49.23\tiny$\pm$10.93 & \underline{54.77\tiny$\pm$6.29}
    %   &81.89\tiny$\pm$2.59 & 84.62\tiny$\pm$2.25 & 70.76\tiny$\pm$5.77 & 58.98\tiny$\pm$2.33 \\
     
    %             Gemma2 & 74.74\tiny$\pm$0.63
    %             &59.58\tiny$\pm$0.50 & 
    %     19.49\tiny$\pm$2.64 & 85.17\tiny$\pm$5.52 
    %   &73.76\tiny$\pm$2.30 & 84.51\tiny$\pm$1.98 & 49.23\tiny$\pm$10.93 & \underline{54.77\tiny$\pm$6.29}
    %   &81.89\tiny$\pm$2.59 & 84.62\tiny$\pm$2.25 & 70.76\tiny$\pm$5.77 & 58.98\tiny$\pm$2.33 \\

    % Chatgpt & 77.83\tiny$\pm$0.63
    %     &65.92\tiny$\pm$0.50 & 
    %     23.20\tiny$\pm$2.64 & 
    %     79.44\tiny$\pm$5.52 
    %   &73.76\tiny$\pm$2.30 & 84.51\tiny$\pm$1.98 & 49.23\tiny$\pm$10.93 & \underline{54.77\tiny$\pm$6.29}
    %   &81.89\tiny$\pm$2.59 & 84.62\tiny$\pm$2.25 & 70.76\tiny$\pm$5.77 & 58.98\tiny$\pm$2.33 \\
        
      \midrule

      \textbf{\sysname{}}(llama) & 88.14\tiny$\pm$0.24 & \textbf{94.30}\tiny$\pm$\textbf{0.78} & \textbf{79.22}\tiny$\pm$\textbf{3.24} & \underline{28.61\tiny$\pm$3.14}    
      & 76.65\tiny$\pm$0.58 & \textbf{88.09}\tiny$\pm$\textbf{0.70} & \underline{62.35\tiny$\pm$5.16} & \textbf{47.51}\tiny$\pm$\textbf{3.11}
      & 92.19\tiny$\pm$0.07 & \textbf{89.36}\tiny$\pm$\textbf{1.07} & \textbf{86.70}\tiny$\pm$\textbf{1.89} & \textbf{51.72}\tiny$\pm$\textbf{3.17}\\

      \textbf{\sysname{}}(qwen)  & \underline{\textbf{88.55}\tiny$\pm$\textbf{0.23}} & 93.20\tiny$\pm$1.12 & 73.18\tiny$\pm$7.03 & \textbf{27.16}\tiny$\pm$\textbf{3.24}
      & \textbf{77.54}\tiny$\pm$0.58 & 87.65\tiny$\pm$1.51 & \textbf{62.46}\tiny$\pm$\textbf{3.35} & 58.58\tiny$\pm$5.39
        & 91.81\tiny$\pm$0.20 & 87.41\tiny$\pm$0.48 & 83.28\tiny$\pm$1.32 & 58.44\tiny$\pm$0.93\\ 

      \textbf{\sysname{}}(gemma)  & 88.49\tiny$\pm$0.42 & 93.43\tiny$\pm$0.53 & \underline{75.06\tiny$\pm$3.29} & 29.68\tiny$\pm$4.72
      & 76.27\tiny$\pm$0.37 & \underline{87.86\tiny$\pm$0.84} & 60.01\tiny$\pm$4.41 & 55.25\tiny$\pm$2.96
        & 91.83\tiny$\pm$0.24 & 87.20\tiny$\pm$0.58 & 82.96\tiny$\pm$0.72 & 57.80\tiny$\pm$1.35\\ 

      \textbf{\sysname{}}(chatgpt)  & 87.48\tiny$\pm$0.70 & \underline{93.61\tiny$\pm$0.50} & 73.30\tiny$\pm$3.16 & 30.42\tiny$\pm$3.68
      & 76.52\tiny$\pm$\textbf{1.02} & 87.17\tiny$\pm$1.39 & 61.09\tiny$\pm$5.46 & 55.84\tiny$\pm$3.00
      & 92.01\tiny$\pm$0.17 & \underline{87.90\tiny$\pm$0.94} & \underline{84.56\tiny$\pm$1.03} & \underline{57.01\tiny$\pm$3.86}\\
    
      \bottomrule
\end{tabular}

\end{threeparttable}
}% end resizebox
\caption{OOD detection performance in terms of Ind-Acc, AUROC, AUPR, and FPR95 (mean $\pm$ std). Best results are bolded, second best are marked with a dash, and $\uparrow$ ($\downarrow$) indicates that larger (smaller) values are better.}
\label{tab:mainresult}
\end{table*}

\subsection{Experimental Settings}

\textbf{Dataset.}
We selected six datasets for the experiments: the citation datasets\texttt{ Cora}, \texttt{Citeseer}, \texttt{Pubmed} \cite{glbench}, and \texttt{Arxiv-2023} \cite{ENGINE}, the web link dataset \texttt{Wikics} \cite{glbench}, and the transaction network dataset \texttt{Photo} \cite{TAGbenchmark}. Since there were no existing benchmarks for text graph OOD detection, in order to evaluate the effectiveness of semantic OOD detection at the node level in text graphs, we primarily construct OOD samples through label shifting. Specifically, we treat one or more classes of samples from the graph as OOD samples, masking them in the training set and treating them as OOD samples in the test set. %The detailed information of the dataset and the OOD class partitioning method can be found in Appendix A.

\textbf{Evaluation Metrics.}
We use AUROC, AUPR, and FPR95 as evaluation metrics for OOD detection, along with accuracy for assessing the classification performance of IND nodes. AUROC evaluates the model’s ability to distinguish IND from OOD samples. AUPR emphasizes the balance between precision and recall, addressing class imbalance. FPR95 measures the misclassification rate of OOD samples as IND at a 95\% IND true positive rate.

% \noindent\textbf{Task.}
% Our primary task is to address the OOD detection problem at the node level in graphs. We aim to detect out-of-distribution nodes in the test set while maintaining the classification performance of in-distribution nodes.

\textbf{Baselines.}
We compare two types of OOD detection models. The first type consists of widely used OOD detection models in the visual domain, such as \textbf{MSP} \cite{msp} and \textbf{ODIN} \cite{odin}. In deployment, we replace the convolutional neural networks with GCNs that extract graph representations. The second type includes advanced baseline models specifically designed for node-level OOD detection in graph data, including \textbf{GPN} \cite{gpn}, \textbf{OOD-GAT} \cite{oodgat}, \textbf{OSSNC} \cite{ossnc} , \textbf{EMP} \cite{Emp} , \textbf{GNNSAFE} \cite{graphsafe}, \textbf{GRASP}\cite{GRASP} and \textbf{NODESAFE}\cite{NodeSafe}. %The detailed description of all baselines can be found in Appendix B.

\textbf{Experimental Setup.}
We implement both our model and all baseline models using PyTorch \cite{pytorch}. Each method utilizes MiniLM \cite{minilm} for text feature extraction and a two-layer GCN to extract node representations, with a hidden layer dimension of 64. A batch normalization layer is applied between the two layers, and the dropout rate is set to 0.5. The Adam optimizer \cite{adam} is used with a learning rate of 0.001 and weight decay regularization set to 0.0005. All models are trained for 300 epochs. For our model, we employ LLAMA3 (8b) \cite{llama3}, Qwen2.5 (7b) \cite{qwen2}, Gemma2 (9b) \cite{gemma}, and Chatgpt-4omini \cite{gpt4} to generate near-OOD and far-OOD data during the large model generation phase. In the contrastive learning phase, the projection layer dimension is set to 128. All experiments are conducted on an Nvidia 3090 GPU.
% Hyperparameters for our model are adjusted based on the in-distribution accuracy on the validation set. For baseline models, hyperparameters are tuned according to the reported results in the respective papers and the validation set performance.

\subsection{Overall Performance}
We conduct experiments on six real-world graph datasets. Due to space constraints, we show three representative datasets (\texttt{Cora}, \texttt{Citeseer}, and \texttt{Pubmed}) in Table \ref{tab:mainresult}. For each dataset, we perform five repeated experiments and report the average results along with the standard deviations for ind-accuracy, AUROC, AUPR, and FPR95. For our method, we select four representative large models to generate pseudo-OOD data.

\begin{figure*}[htbp] % 使用 figure* 来跨越两栏
    \centering
    \includegraphics[width=0.93\textwidth]{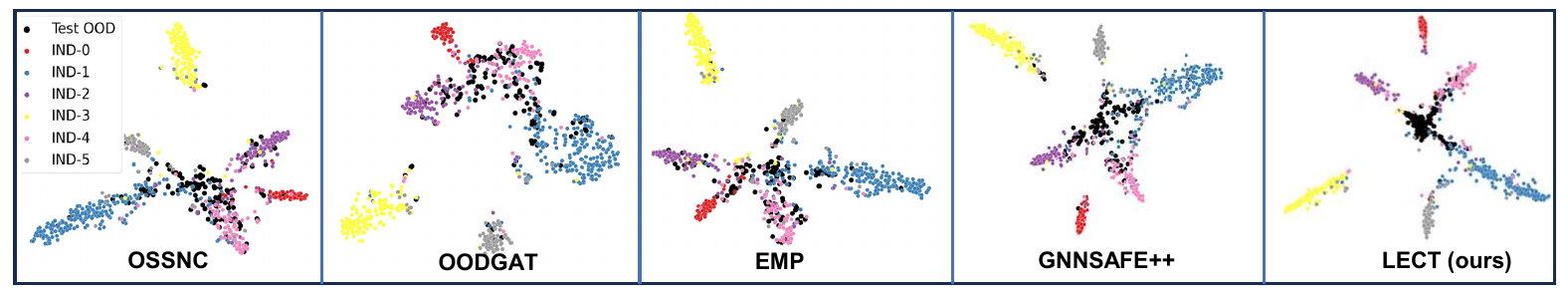} % 使用页面宽度填充
    \caption{ t-SNE visualization of node embeddings on the \texttt{Cora} for different baseline models and \sysname{}.}
    \label{fig:cora_baselines}
\end{figure*}

As shown in Table \ref{tab:mainresult}, our proposed method outperforms all baseline models on all datasets in OOD detection. Specifically, on the \texttt{Cora}, the AUROC is improved by 1.3\% compared to the best baseline model, AUPR is improved by 7.9\%, and FPR95 is reduced by 2.1\%. On the \texttt{Citeseer}, AUROC is improved by 3.6\%, AUPR is improved by 11.2\%, and FPR95 is reduced by 7.2\%. On the \texttt{Pubmed}, all metrics show a performance improvement ranging from 4.7\% to 11\%. These results demonstrate the effectiveness of our proposed method, which combines LLM-generated pseudo-samples with energy-based contrastive learning for OOD detection. Additionally, we observe that, compared to other methods, the accuracy of our IND samples is even improved, with only a slight performance loss on the \texttt{Pubmed}. This indicates that our method can accurately detect OOD samples while maintaining strong node classification performance. As shown in Figure \ref{fig:cora_baselines}, LECT outperforms baseline models by maintaining high accuracy in classifying IND nodes and significantly improving the distinction of OOD samples.

\begin{figure*}[htbp] % 使用 figure* 来跨越两栏
    \centering
    \includegraphics[width=0.93\textwidth]{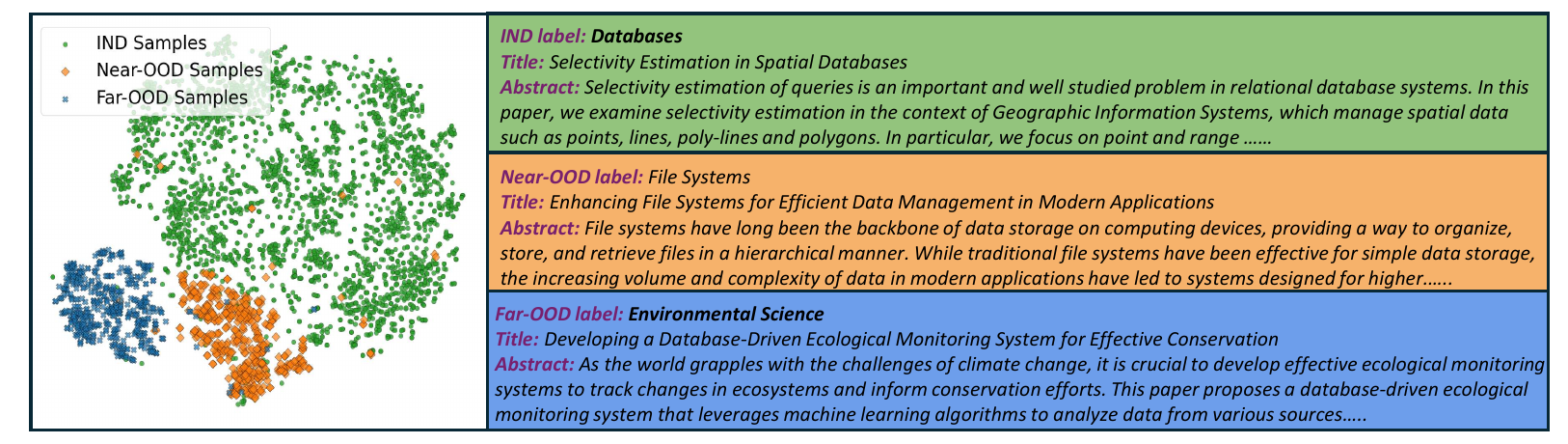} % 使用页面宽度填充
    \caption{t-SNE visualization and textual representation of the generated near-OOD and far-OOD samples on the \texttt{Citeseer}.}
    \label{fig:casestudy}
\end{figure*}

% Furthermore, we observe that the GNNSAFE(++) model is the best-performing baseline, highlighting the superiority of energy-based methods for OOD detection. Our method demonstrates substantial improvements over GNNSAFE, primarily because, in addition to generating high-quality samples using LLM, we incorporate an energy-based contrastive learning approach. In contrast to GNNSAFE++, which imposes energy bounds using thresholds, potentially leading to significant drops in ind-Acc and suboptimal OOD detection performance, our method more effectively captures the dependency relationship between IND and OOD samples. This results in enhanced detection capabilities while obviating the need for arbitrary threshold settings.
Our method achieves notable performance improvements by integrating high-quality pseudo-OOD sample generation using LLMs with an energy-based contrastive learning framework. This joint design facilitates more effective modeling of the relationship between IND and OOD samples, improving detection accuracy without the need for manually tuned thresholds. Unlike prior energy-based approaches that rely on static bounds and often compromise IND classification performance, our method maintains a better balance between IND classification and OOD detection. Specifically, we leverage four LLMs to generate pseudo-OOD node attributes that exhibit both distributional and semantic shifts from the IND data, as illustrated in Figure~\ref{fig:casestudy}. %A comparison with direct LLM-based detection, provided in Appendix~C, shows that our LECT model consistently outperforms the baseline.

\begin{figure}[t!]
    \centering
    \includegraphics[width=0.45\textwidth,scale=1.00]{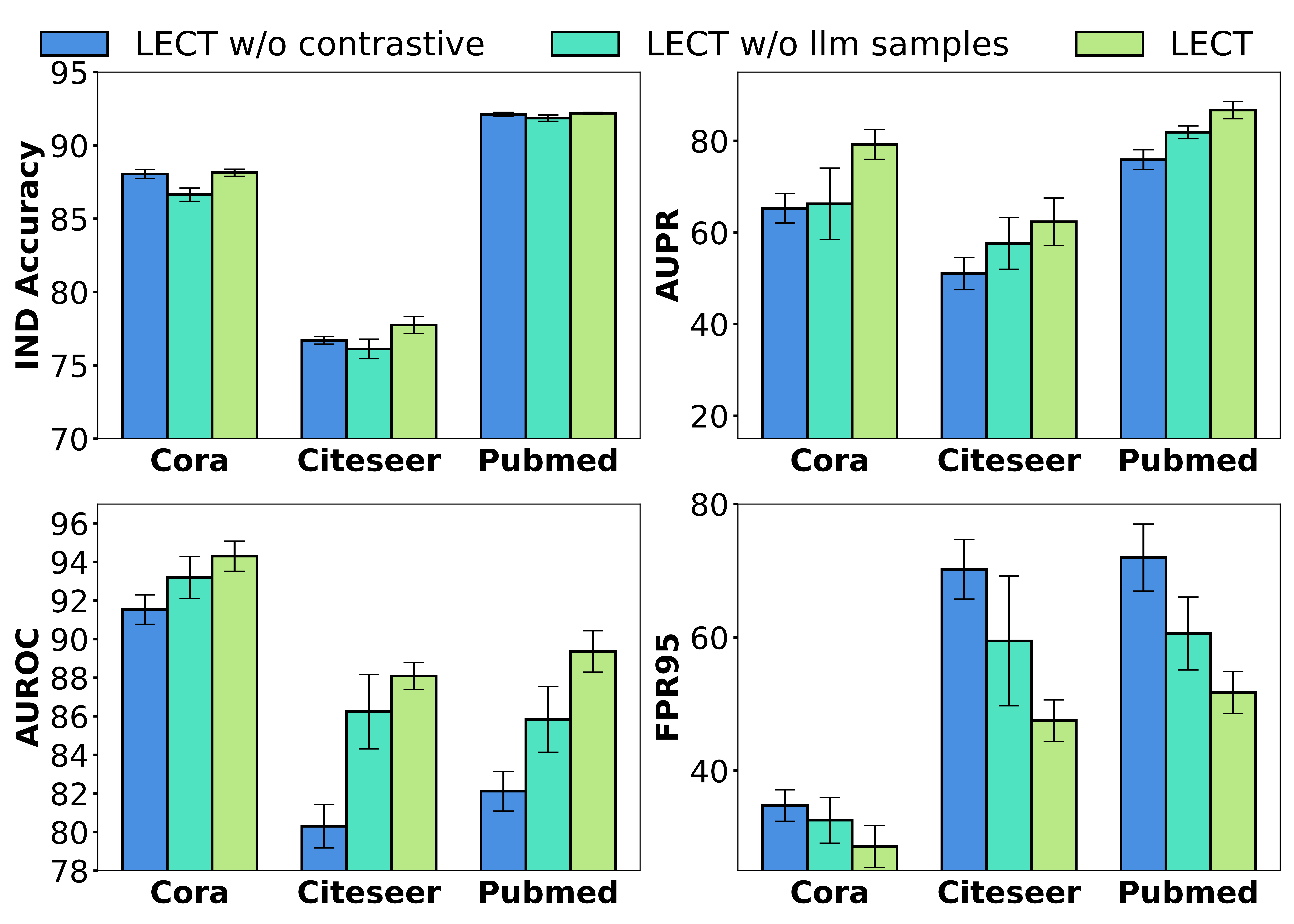}
    \caption{Ablation study results of \sysname{} on the \texttt{Cora}, \texttt{Citeseer}, and \texttt{Pubmed} datasets, showing the performance without contrastive learning and without LLM-generated samples, respectively.}
    % \vspace{-5pt}
    \label{fig:ablation_llm}
  \end{figure}

\begin{figure}[htbp]
    \centering
    \includegraphics[width=0.45\textwidth,scale=1.00]{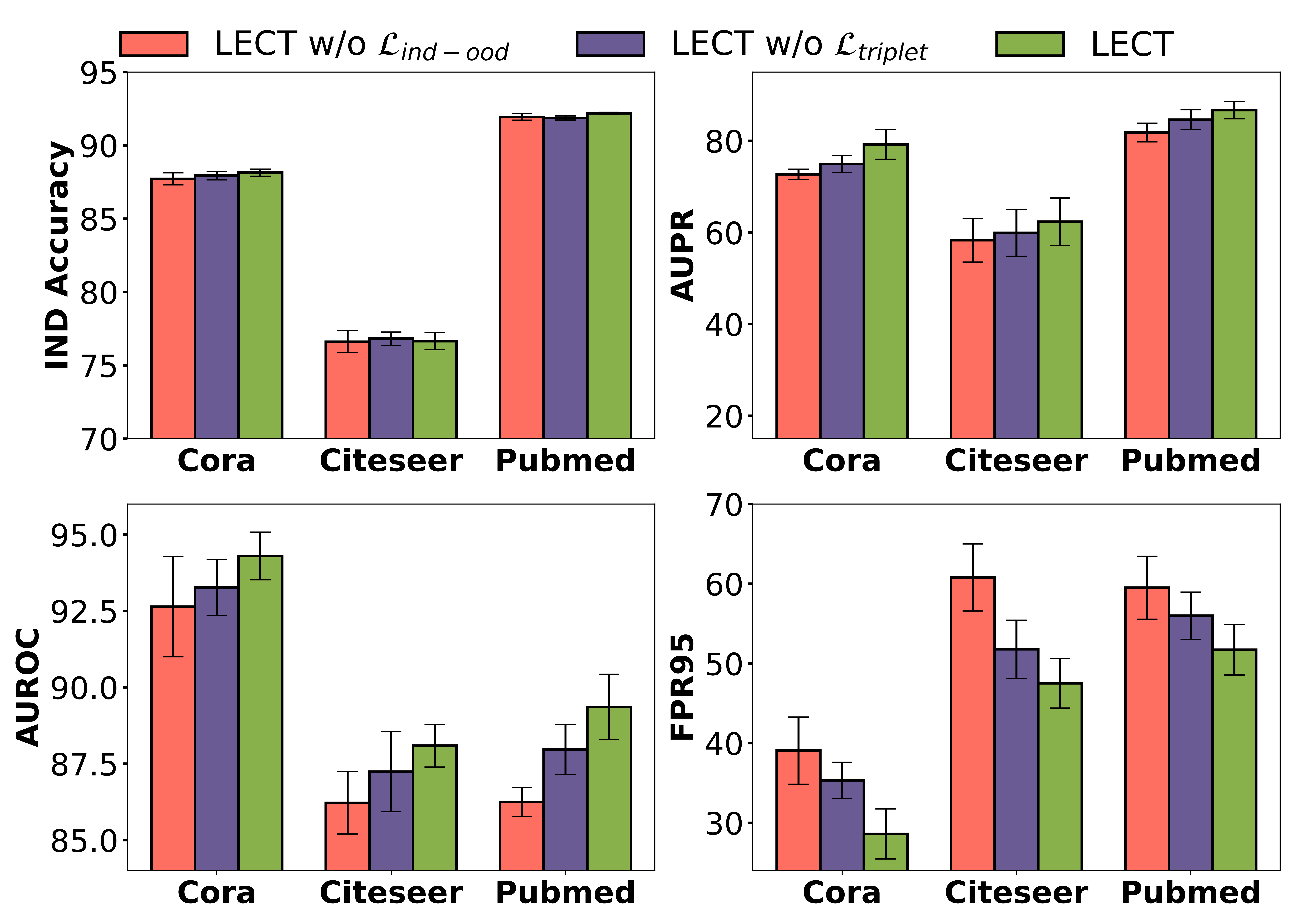}
    \caption{Ablation study results of \sysname{} on the \texttt{Cora}, \texttt{Citeseer}, and \texttt{Pubmed} datasets, showing the performance without $\mathcal{L}_{\text{ind-ood}}$ and without $\mathcal{L}_{\text{triplet}}$, respectively.}
    % \vspace{-5pt}
    \label{fig:ablation_l_indood}
  \end{figure}

\subsection{Ablation Study}

\textbf{Ablation Study On the Role of LLM-Generated Pseudo-OOD Samples and Contrastive Learning.}
To investigate the contributions of LLM-generated pseudo-OOD samples and contrastive learning in OOD detection, we perform ablation experiments by removing these components individually: (i) LECT \textit{w/o contrastive learning} excludes the proposed contrastive learning method, relying solely on energy scores for OOD detection without using OOD samples and contrastive learning; (ii) LECT \textit{w/o LLM-generated samples} replaces the LLM-generated pseudo-OOD samples with randomly generated text attributes from other datasets. %The specific procedure is described in Appendix C.
The experimental results are presented in Figure \ref{fig:ablation_llm}. \textbf{(i) Removal of Contrastive Learning:} When only the energy score is used for OOD detection, the performance is significantly worse compared to the case where pseudo-OOD samples generated from random text are combined with contrastive learning. This demonstrates that incorporating pseudo-OOD samples and contrastive learning significantly enhances OOD detection performance. \textbf{(ii) Removal of LLM-generated Pseudo-OOD Samples:} We observe that using randomly generated text attributes results in poorer performance compared to using LLM-generated attributes. This highlights the effectiveness of our COT template in guiding large models to generate high-quality pseudo-OOD data with meaningful dependencies.

\textbf{Ablation Study on the Role of $\mathcal{L}_{\text{ind-ood}}$ and $\mathcal{L}_{\text{triplet}}$.}
To better explore the roles of $\mathcal{L}_{\text{ind-ood}}$ and $\mathcal{L}_{\text{triplet}}$, we conducted ablation experiments, with the results shown in Figure~\ref{fig:ablation_l_indood}.  \textbf{(i) Removal of $\mathcal{L}_{\text{ind-ood}}$:} When training with only $\mathcal{L}_{\text{triplet}}$ and $\mathcal{L}_{\text{sup}}$, the overall performance is inferior to that achieved with $\mathcal{L}_{\text{ind-ood}}$ and $\mathcal{L}_{\text{sup}}$. This indicates that the Linked IND-OOD Pairs play a critical role in the overall model performance. This is mainly because the LLM-generated samples are inherently dependent on the linked IND samples, and this loss directly captures such dependencies.\textbf{(ii) Removal of $\mathcal{L}_{\text{triplet}}$:} When training with only $\mathcal{L}_{\text{ind-ood}}$ and $\mathcal{L}_{\text{sup}}$, the performance is inferior to the joint training with $\mathcal{L}_{\text{ind-ood}}$, $\mathcal{L}_{\text{triplet}}$, and $\mathcal{L}_{\text{sup}}$. This highlights the necessity of the designed Triplet Contrastive Pairs, as this loss captures the differences between normal connections (IND-OOD) and OOD connections (IND-OOD) in the graph. Through contrastive learning, it enhances the energy gap between OOD samples and ind samples.

%In addition, we perform ablation experiments to analyze different training paradigms for pre-trained language models, with the results provided in Appendix C.
\subsection{Sensitivity Analysis}

\begin{figure}[t!] 
    \centering
    \begin{subfigure}{0.23\textwidth}
        \centering
        \includegraphics[width=\linewidth]{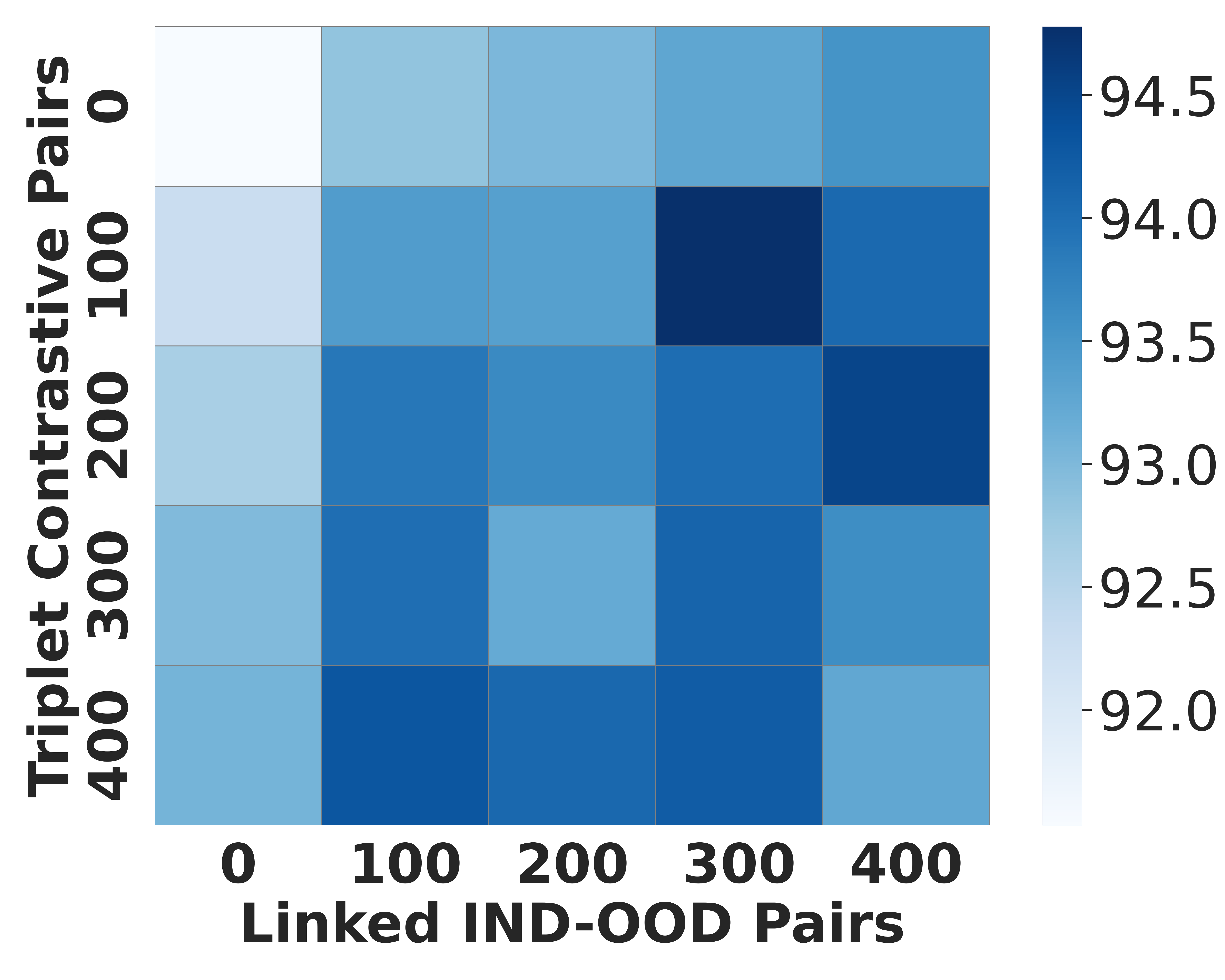}
        \caption{\texttt{Cora}}
        \label{fig:cluster_acc}
    \end{subfigure}%
    \begin{subfigure}{0.23\textwidth}
        \centering
        \includegraphics[width=\linewidth]{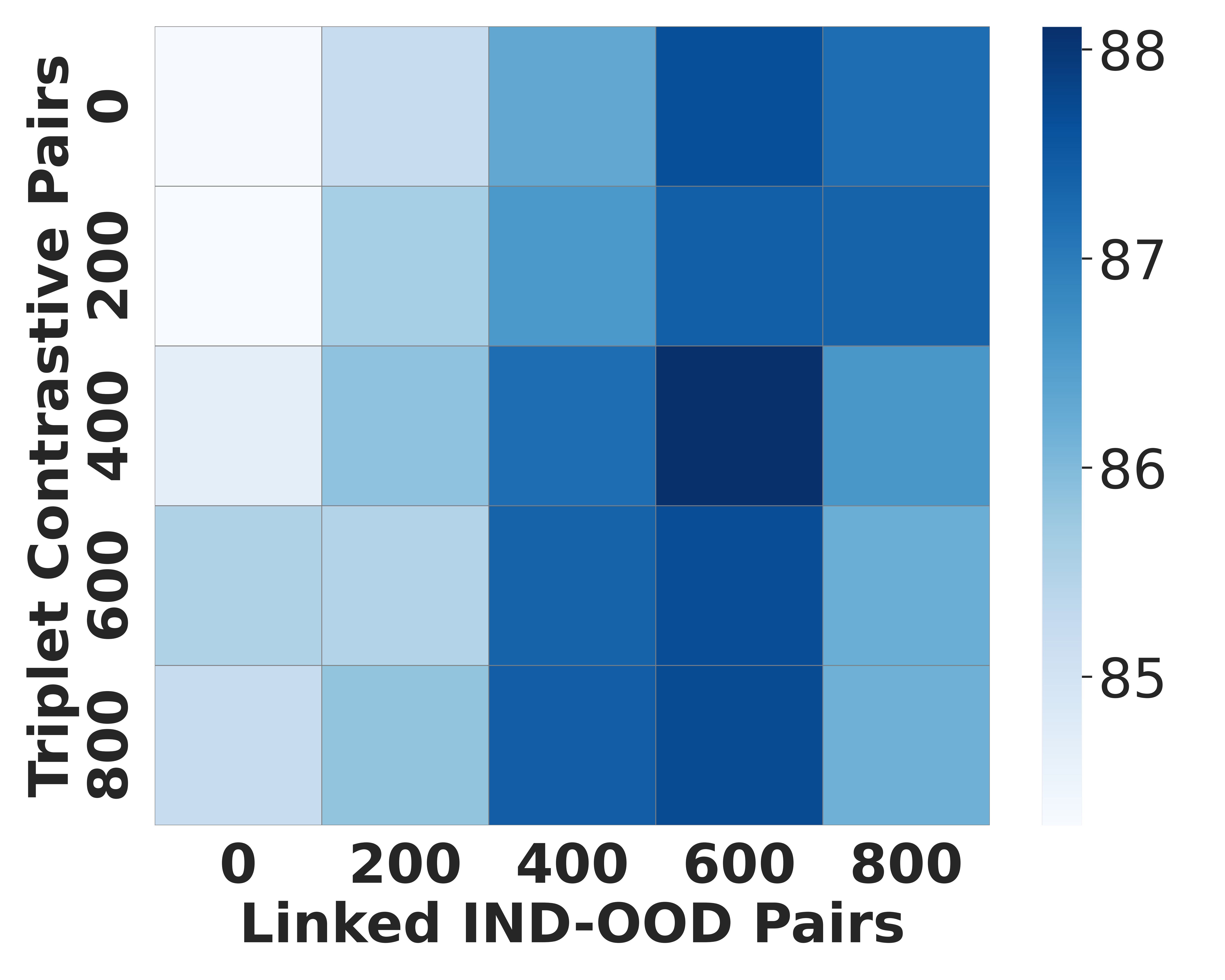}
        \caption{\texttt{Citeseer}}
        \label{fig:cluster_auc}
    \end{subfigure}
    \caption{Heatmap of AUROC values for different sampling sizes of Linked IND-OOD pairs and Triplet Contrastive Pairs.}
    \label{fig:heatmap}
    % \vspace{-3mm}
\end{figure}

To evaluate the impact of Linked IND-OOD pairs and Triplet Contrastive Pairs in contrastive learning, we sampled different quantities of sample pairs, and the results are presented in Figure \ref{fig:heatmap}. It is observed that the model performs worst when the sample sizes for both types of pairs are zero, indicating that both sampling strategies contribute positively to the model’s ability to learn OOD patterns. For the \texttt{Cora} dataset, the optimal performance is achieved with 300 and 100 sample pairs, while for the \texttt{Citeseer} dataset, the best performance occurs with 600 and 400 sample pairs, respectively. This suggests that the number of sampled pairs influences the results across different datasets. In general, the number of sampled pairs is positively correlated with the size of the dataset. However, we also observe that an excessive number of samples does not always lead to improved performance, as it may result in overfitting, ultimately reducing the model’s effectiveness. %In addition, we conducted experiments on the impact of different text encoders, graph encoders, and the hyperparameter $\gamma$ on the overall performance, with detailed results provided in Appendix~C.

\section{Conclusion}
In this paper, we address the critical challenge of OOD detection in text-attributed graphs, proposing a novel method that integrates LLM with contrastive learning. Our approach improves OOD detection by generating high-quality pseudo-OOD samples that preserve dependency relationships between nodes. We introduce an energy-based contrastive learning framework that effectively distinguishes between IND and OOD nodes while maintaining robust node classification performance. Experimental results demonstrate that our method outperforms existing state-of-the-art approaches across six benchmark datasets, highlighting its effectiveness in both OOD detection and node classification tasks.

\section{Acknowledgments}
This work done by Xiaoxu Ma and Minglai Shao is supported by the National Natural Science Foundation of China (No. 62272338) and the Research Fund of the Key Lab of Education Blockchain and Intelligent Technology, Ministry of Education (EBME25-F-06). 
Dong Li, Xintao Wu, and Chen Zhao did not receive any financial support for this work and contributed only by developing the research ideas, participating in discussions, and providing feedback on the manuscript.

\bibliography{aaai2026}
% \section{Reproducibility Checklist}
%     \label{sec:ReproducibilityChecklist}
%     \input{ReproducibilityChecklist}
\clearpage
% \appendix
%     \label{sec:appendix}
%     \input{appendix}

\end{document}